\documentclass[runningheads]{llncs}
\usepackage{graphicx}
\usepackage{tabularx}
\usepackage{caption}
\usepackage{booktabs}
\usepackage{diagbox}
\usepackage{array}
\usepackage{multicol}

\usepackage{appendix}
\usepackage{todonotes}
\usepackage{standalone}
\usepackage{textcomp,mathcomp}
\usepackage{wasysym}
\usepackage{xcolor}
	
\usepackage[euler]{textgreek}

\usepackage{todonotes}

\newcommand{\dbo}[1]{{\color{black} #1}}
\newcommand{\jde}[1]{{\color{black} #1}}
\newcommand{\wri}[1]{{\color{black} #1}}

\newcommand{\fgh}[1]{{\color{black} #1}}

\begin{document}
\title{Enhancing Multi-Objective Optimization through Machine Learning-Supported Multiphysics Simulation
}
\titlerunning{MO-Optimisation trough ML-Supported Multiphysics Simulation}

 \author{Diego Botache\inst{1}\orcidID{0000-0003-1694-0307} \and
 Jens Decke\inst{1}\orcidID{0000-0002-7893-1564} \and \\
 Winfried Ripken\inst{2}\orcidID{0009-0006-1926-6695} \and
 Abhinay Dornipati\inst{3} \and 
 Franz Götz-Hahn\inst{1}\orcidID{0000-0003-3465-5040} \and \\
 \hspace{-0.25cm}Mohammed Ayeb\inst{3}\orcidID{0000-0001-9479-059X}\and Bernhard Sick\inst{1}\orcidID{0000-0001-9467-656X} 
 }
 \authorrunning{D. Botache et al.}
 %
 \institute{Intelligent Embedded Systems, University of Kassel, Hessen, Germany 
 \email{\{diego.botache, jens.decke, franz.goetz.hahn, bsick\}@uni-kassel.de}
 \url{https://www.uni-kassel.de/eecs/ies}
 \and
 Technische Universität Berlin, Berlin, Germany \\
 \email{winfried.ripken@gmail.com}\\
 \url{https://web.ml.tu-berlin.de/}
 \and
Vehicle Systems and Fundamentals of Electrical Engineering,\\ University of Kassel, Hessen, Germany \\
 \email{\{abhinay.dornipati, ayeb\}@uni-kassel.de}\\
 \url{https://www.uni-kassel.de/eecs/fsg}
 }

\maketitle              
\begin{abstract}

This paper presents a methodological framework for training, self-optimising, and self-organising surrogate models to approximate and speed up multiobjective optimisation of technical systems based on multiphysics simulations. At the hand of two real-world datasets, we illustrate that surrogate models can be trained on relatively small amounts of data to approximate the underlying simulations accurately. Including explainable AI techniques allow for highlighting feature relevancy or dependencies and supporting the possible extension of the used datasets. One of the datasets was created for this paper and is made publicly available for the broader scientific community. Extensive experiments combine four machine learning and deep learning algorithms with an evolutionary optimisation algorithm. The performance of the combined training and optimisation pipeline is evaluated by verifying the generated Pareto-optimal results using the ground truth simulations. The results from our pipeline and a comprehensive evaluation strategy show the potential for efficiently acquiring solution candidates in multiobjective optimisation tasks by reducing the number of simulations and conserving a higher prediction accuracy, i.e., with a MAPE score under 5\% for one of the presented use cases.

\end{abstract}

\hspace{1cm}

\section{Introduction}
\label{sec:intro}
\dbo{
Multiphysics and multiscale simulations have become crucial for the computational modelling and analysis of multiple interacting physical phenomena in technical systems. These phenomena include mechanics, fluid dynamics, heat transfer, and electromagnetics for \fgh{a wide variety of} applications, such as aerospace engineering, biomedical engineering, and materials science\fgh{, to name a few}. 
\dbo{Incorporating multiple physical phenomena into simulations is a powerful tool for engineers, enabling them to investigate various design alternatives and parameters and enhance the depth of their decision-making during design processes. Typically, this approach involves considering competing objectives simultaneously within multiobjective optimization tasks, thereby ensuring that the final design solutions strike an optimal balance across diverse performance criteria.}}

\begin{figure}[b!]
    \centering
    \includegraphics[width=.8\linewidth]{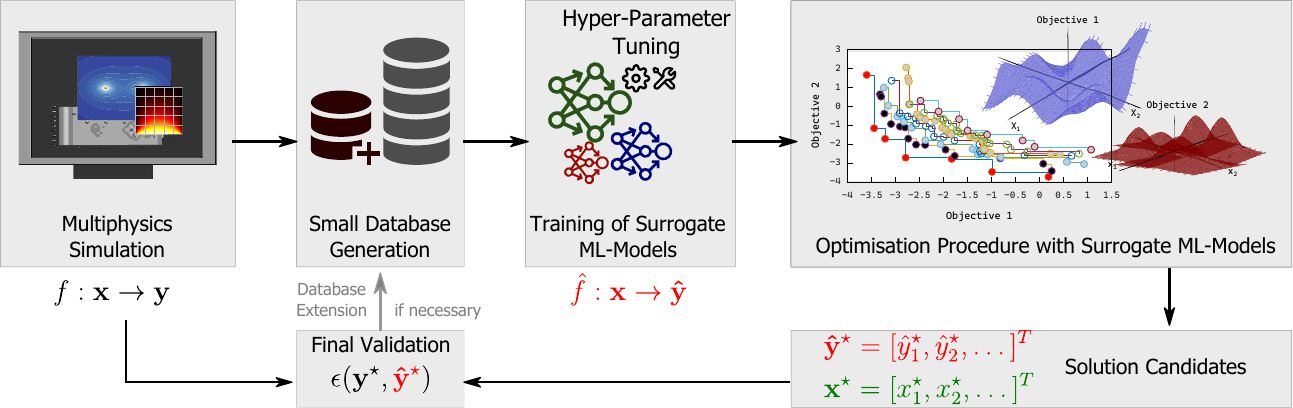}
    \caption{Proposed strategy for training and self-optimising surrogate models using machine learning and deep learning techniques to tackle multiobjective optimisation problems in complex multiphysics simulations.}
    \label{fig:intro}
\end{figure}

\dbo{Acquiring optimal solutions presents a significant challenge due to the complex nature of numerical models and potential nonlinear dependencies among design parameters. Moreover, constrained solution spaces yield scarce feasible solutions and require the inclusion of advanced domain knowledge of the underlying physical problem. To address these issues, we propose using surrogate models, i.e. data-driven algorithms, as an alternative to running computationally expensive multiphysics simulations in combination with advanced optimisation techniques i.e. evolutionary algorithms. We demonstrate how surrogate models in multiphysics simulations reduce computational burdens, accelerating the design process and enabling broader exploration of design options. 

As shown in Fig.~\ref{fig:intro}, our strategy can be summarised as follows: First, we identify the multiphysics system input and output space followed by a small data generation procedure. Next, we apply our pipeline for the training and evaluation of machine learning (ML) models, which are then used for the optimisation task of the input parameters of the multiphysics system. Finally, we validate the optimisation results by comparing the outputs of the ML models against the simulated values of selected solution candidates of the multiobjective optimisation problem.}

\wri{
The main contributions of our paper can be summarised as follows:
\begin{enumerate}
\item A real-world tabular dataset using multiphysics simulations, which is made publicly available
\item We streamline the efficient acquisition of solution candidates in multiobjective optimisation tasks using multiple surrogate models in combination with evolutionary algorithms
\item We validate the acquired results and derive insights about the relevance and dependencies of features in our real-world examples by using explainable AI techniques
\end{enumerate}}

\dbo{The remainder of this article is structured as follows: Section~\ref{sec:related_work} summarises the related work, followed by a detailed description of our combined training and optimisation pipeline in Section~\ref{sec:framework}, Section~\ref{sec:use_cases} describes the underlying physical problems of the two real-world tabular datasets used in our approach. The results of our evaluation strategy for training and optimisation with an examination of the acquired results with explainable AI methods are presented in Section~\ref{sec:results}.} \fgh{Finally, Section~\ref{sec:conclusion} summarises the main conclusions and open issues for future work.}

\section{Related Work}
\label{sec:related_work}

\dbo{Coupling multiphysics simulations to model a system may compromise stability, accuracy, and robustness~\cite{groen_survey_2014,multiPhy13}. Researchers employ data-driven approaches, utilizing machine learning (ML) and deep learning techniques to predict efficiently and approximate multiphysics processes, where deep learning models typically require large amounts of data for training and feature extraction~\cite{de_bezenac_deep_2019}.
\fgh{While individual approaches to substituting simulations with ML techniques have been proposed, t}here is an absence of a general strategy for training, hyperparameter tuning, evaluating and comparing diverse data-driven models in multiphysics simulations. Evaluations are often customised for specific applications and \fgh{it is unclear, whether the approach can be generalized to other problems} and especially in cases with sparse data. 
}

\subsection{Surrogate Modelling}
\dbo{
Common approaches involve using submodels to represent distinct physical processes and coupling them to determine the contribution to the overall complex physical system~\cite{groen_survey_2014}. When dealing with tabular data, employing boosted trees or ensemble strategies for meta-modelling is a reliable benchmark for execution speed, computational efficiency, and accuracy~\cite{derouiche_data-driven_2021}. In our research, we depict the multiphysics problem in tabular format, enabling direct training of surrogates for predicting individual output values of the associated physical system. This approach allows hyperparameter tuning of each surrogate and gives insight into the models' performance for each system output.

Besides fixed tabular or parameterised physical problems, integrating deep learning techniques with numerical simulation tools has shown the potential to reduce computational burden using more complex data. These techniques can predict multiple outputs of a physical system and extract in-depth features of the data\fgh{. For example, convolutional auto-encoders have been used to model} scalar transport equations coupled to Navier–Stokes equations~\cite{el_haber_deep_2022}. Further application strategies in fluid flow applications~\cite{morimoto_convolutional_2021,bhatnagar_prediction_2019}, heat transfer~\cite{sharma_review_2023,hachem_deep_2021} and electromagnetics~\cite{PFS22} show the potential for accurate modelling of multiphysics problems but also on predicting \fgh{problem specific} key performance indicators. Still, most of these approaches lack the explainability, interpretability of \fgh{the trained or }extracted features and require high amounts of data. Physics-informed neural networks~\cite{sharma_review_2023,raissi_physics-informed_2019,karniadakis_physics-informed_2021} \fgh{have been proposed as} a powerful tool to obtain solutions for multiphysics simulations without access to ground-truth data. However, these models require substantial knowledge of the underlying system, making them less accessible to non-experts. Additionally, they can be computationally expensive and time-consuming to train, needing significant resources to converge to accurate solutions. 

}

\subsection{Multiphysics Optimisation and Data Extension}
\dbo{In many cases, the goal consists of optimising the design parameters within these technical systems, considering multiobjective criteria.} \wri{Multiobjective optimisation requires finding trade-offs between potentially conflicting objectives~\cite{sener2018multi}, which quickly becomes a challenging task for automated optimisation algorithms. A recent study investigated this problem through the lens of diversity and showed that their approach increases diversity without sacrificing global performance~\cite{pierrot2022multi}. In our work, we shift the decision to consider the different objectives of the user by presenting a range of Pareto-fronts of optimised designs. We additionally encourage diversity in our optimised results by carefully choosing combinations of surrogate models and optimisation algorithms.}

\dbo{Limited data also poses significant challenges in multiphysics optimisation, and only a few works highlight the challenges in specific use cases. An investigation integrates topology optimisation and generative models (e.g., generative adversarial networks) in a framework allowing the exploration of new design options, thus generating many designs starting from limited previous design data~\cite{OJK+19}. 
\jde{In a preliminary investigation of one of the use cases, novelty and anomaly detection algorithms were used in design optimisation tasks. The study found that these algorithms are effective in exploring the design space, but they have limitations when it comes to exploitation~\cite{DSB+22}. To address this disadvantage, deep active design optimization was introduced combining the fields of deep active learning and design optimization~\cite{decke2023dado}.}}
\section{Machine Learning Supported Optimisation Strategy}
\label{sec:framework}

\begin{figure}[b!]
\centering
\includegraphics[width=\textwidth]{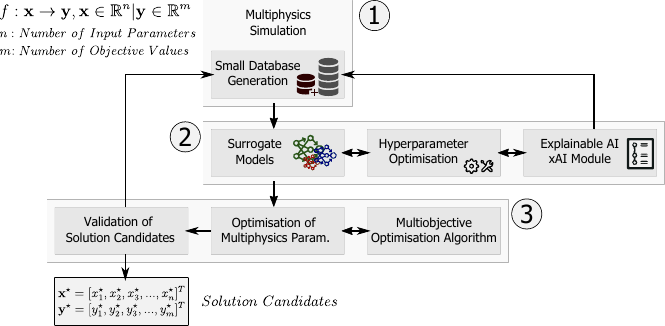}
\caption{Ilustration of our proposed Pipeline for self-optimising surrogate models, which can be seamlessly integrated into the optimisation process of multiphysics problems and comprises three main blocks: data acquisition, surrogate model training, and multiobjective optimisation.}
\label{fig:pipeline}
\end{figure}

\dbo{
In this section, we present our pipeline for applying, explaining and evaluating ML surrogate models coupled with multiobjective optimisation strategies of technical systems. It streamlines the evaluation and experimentation process and presents a general validation and performance evaluation procedure.
Our approach, depicted in Fig.~\ref{fig:pipeline}, consists of three main blocks, which will be explained separately in the following subsections.}

\subsection{Data Acquisition}
\label{subsec:data_acquisition}
\dbo{The first block \textbf{(1)}, shown in Fig.~\ref{fig:pipeline}, consists of the data acquisition module, \fgh{which, in the initial phase, is a manual process.} Here, a small database of simulation results from available numerical models \fgh{are gathered, functioning as a starting point of the optimisation pipeline}. This step \fgh{is crucial in the success of the optimisation process and} poses specific requirements \fgh{and considerations. On the one hand, the initial database must cover many feasible input parameter combinations to train the surrogate models efficiently. On the other hand, the generation process itself relies on slow and complex numerical models, implicitly restricting the size of the database.}
\fgh{Naturally, the quality of this first selection of data impacts} the accuracy and reliability of the subsequently trained machine-learning models and the outcome of the optimisation process as a whole.
In the context of the outlined use cases, the supplied tabular data comprises singular simulation instances. It can be extended to non-parametric or tabular approaches like pictures or videos but this is not the scope of this paper. Consequently, instances characterized by incomplete simulation results or missing data will be systematically filtered from the database.  
In addition, depending on the type of surrogate model, some data pre-processing is essential and can be defined by the pipeline user.
}

\subsection{Surrogate Models}
\label{subsec:surrogate_models}
\dbo{The extreme gradient boosting (XGB)~\cite{chen_xgboost_2016} serve as our baseline surrogate model and we train single models for the prediction of individual output values. In contrast, deep learning techniques are used for predicting multiple target values, therefore they can consider relationships in the output space.
The training of the surrogate models is shown in Fig.~\ref{fig:pipeline} block \textbf{(2)}. We employ hyperparameter optimisation strategies to ensure robust training of the models. Since we assume that the databases are limited and usually distributed in a specific range of parameter values it is still possible to generate solution candidates outside that range at the end of the proposed pipeline, but this could affect the prediction performance of the models. Therefore, to identify possible unexplored areas that should be generated in the first block, Fig.~\ref{fig:pipeline} (1), it is crucial to understand the dependencies present in the data. To achieve this, our approach incorporates explainable artificial intelligence (xAI) techniques that provide useful insights into the data will be explained in detail in the following. }

\subsection{Interpretable Surrogate Modelling (xAI Module)}
\label{subsec:xai}
\dbo{By analysing the results obtained from selected xAI methods, we can identify features or parameters in the data that highly influence the outputs of the physical systems. Additional methods can also provide information about the distribution of the parameter values, showing areas of the data that are not fully explored, and therefore extend the database accordingly.
We used feature relevances from the XGB surrogates to understand the impact of input features on target values. Furthermore, we extended our approach using partial dependence plots~\cite{zhao_causal_2021} to identify whether the relationship between relevant features and target values is linear, non-linear, monotonic, or more complex. These plots are a powerful technique to show the marginal effects of one or two variables on the predicted outcome of a machine learning model while holding all other variables constant. Additional xAI techniques can be included in the pipeline but will not be further considered in this study.}

\subsection{Multiobjective Optimisation and Validation}
\label{subsec:optimisation}
\dbo{The third block \textbf{(3)} of our approach leverages the pre-trained machine learning models to perform the multiobjective optimisation task, including evolutionary algorithms. The optimisation algorithm used in the following use cases is a non-dominated sorting genetic algorithm (NSGA). Genetic algorithms have proven to be a robust and reliable design optimisation method~\cite{CCB+23}. Therefore, we integrated the NSGA-2 into our optimisation pipeline~\cite{Deb2002a}.  
The results provided by the evolutionary algorithm will be evaluated using Pareto frontiers, and the final selection of solution candidates will correspond to the points at the Pareto front. A validation step is required to validate the prediction performance of the surrogate models within the selected solution candidates. After the validation step, we propose using performance indicators~\cite{ishibuchi_modified_2015,fonseca_improved_2006} to compare the simulated Pareto frontiers from multiple experiments with different surrogate models.
}
\section{Experimental Design}
\label{sec:use_cases}
\dbo{
In the following use cases we explain the optimisation task and the corresponding target values to consider for training the surrogate models.
For each target value, we train a single XGB Regressor considering the mean squared error (MSE) and optimise each model's hyperparameters independently using a combined cross-validation and Bayesian optimisation strategy~\cite{balandat_botorch_2020}. The XGB as a baseline provides valuable insights into the underlying problem and data dependencies.  
In addition to the XGB surrogate models, we employ ensemble strategies combining multiple scikit-learn regressors at the decision level~\cite{feurer-neurips15a,scikit-learn}. A two-step process to train and optimise the hyperparameters consists of first performing a random search of hyperparameters to identify promising regressors. Then, we combine these regressors into an ensemble using a weighted average, where the weights are learned through gradient-based optimisation. The ensemble is trained using cross-validation, and we select the best-performing ensemble based on validation scores. The final ensemble is trained on the entire dataset and used for prediction. This approach balances exploring the hyperparameter space and exploiting the most promising models. 
Finally, we use two deep-learning methods, MLP and CNN, for the regression task. We use both models to estimate all target values, and we tune the hyperparameters of each model using a combined cross-validation and Bayesian optimisation strategy. In this study, we focus on evaluating the regression performance using deterministic metrics.
}

\subsection{Use Case 1: Motor Dataset}
\label{subsec:usecase1}
The first use case is concerned with optimising the performance of a particular topology of an electric motor. The objectives of the optimisation problem correspond to maximising the mean \textbf{Torque~$J_M$} while minimising the \textbf{Total Loss~$J_\Phi$} and the \textbf{Magnet Mass~$J_m$}. Therefore, the optimisation objectives are conflicting in nature.
\begin{figure}[b!]
    \begin{minipage}{0.48\textwidth}
    \vspace{0.5cm}
        \centering
        \includegraphics[width=\linewidth]{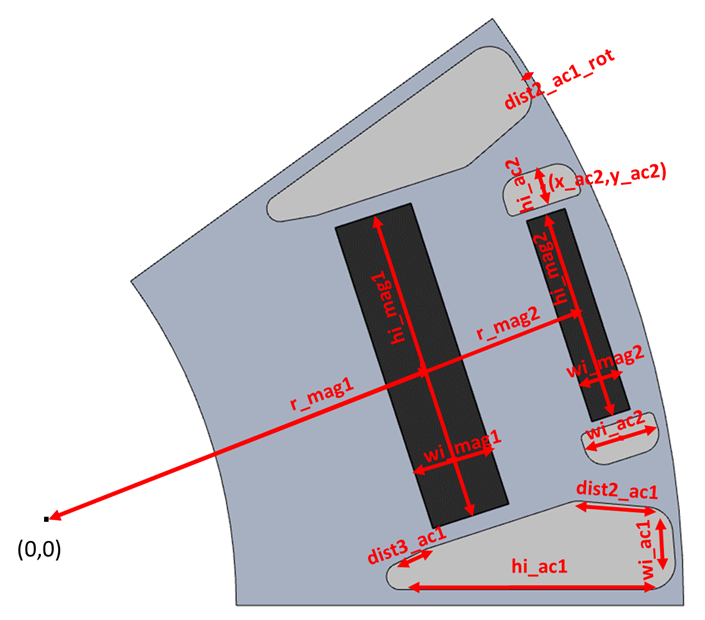}
        \caption{Parameterised 2D rotor segment of the baseline machine with black fields representing the magnet components and grey areas for the air cavities.}
        \label{fig:use_case_1}
    \end{minipage}
    \hfill
    \begin{minipage}{0.48\textwidth}
        \centering
        \captionof{table}{Dimensions, parameters, constraints, and performance characteristics}
        \label{tab:Use_Case1}
        \begin{tabular}{|>{\centering}p{3cm}|>{\centering}p{2cm}|}
             \hline 
             \textbf{Dimension or Parameter}&   \textbf{Value}\tabularnewline 
             \hline
             Stator outer \diameter & 230 [mm]\tabularnewline
             Rotor outer \diameter &   152 [mm] \tabularnewline
             Airgap Length &   1 [mm] \tabularnewline
             Magnet Remanence at 120 $\tccentigrade$ &   1.18 [T] \tabularnewline
             \hline
             \textbf{Constraints}&   \textbf{Value} \tabularnewline
             \hline
             Maximum Current   & 636.34 [A]\tabularnewline
             Maximum Voltage   & 270 [V]\tabularnewline
             \hline
             \textbf{Performance Characteristics}& \textbf{Value}\tabularnewline
             \hline
             $J_M$ @1000 rpm   & 280.5 [Nm]\tabularnewline
             $J_\Phi$ @1000 rpm   & 3.76 [kW]\tabularnewline
             $J_m$   & 2.78 [kg]\tabularnewline
             \hline
        \end{tabular}

    \end{minipage}
\end{figure}
The baseline machine considered for optimisation is described in terms of its dimensions, i.e. parameters, constraints, and performance characteristics in Table \ref{tab:Use_Case1}. For this work, only the rotor topology of the baseline machine---a 60-slot 10-pole internal permanent magnet synchronous machine---is parameterised with 15 geometric parameters. These parameters govern the size and position of the magnets and air cavities, which have a significant impact on all the target values.
As shown in Fig.~\ref{fig:use_case_1}, there are two lateral magnets embedded into the rotor iron lamination. In addition, two air cavities are designed in the rotor topology to guide the magnetic flux from the magnets into the air gap at the stator-rotor interface.  

Efficient sampling of high dimensional spaces is crucial to deliver enough information about the design space and thus for training ML surrogate models. This is usually constrained by the computational resources with the numerical models and exploring every possible combination of parameters becomes unfeasible. Design of experiments and specially Latin Hypercube Design (LHD) sampling limits the number of simulations and enables systematic and efficient exploration across the parameter space of the motor topology. The COMSOL 6.0 uncertainty quantification module generates 691 sample points based on this algorithm i.e., motor design variations. The training and test datasets consist of 552 and 139 designs respectively.

 \begin{figure}[b!]
	\centering
    \rotatebox{-90}{
    \scalebox{0.5}{\includegraphics{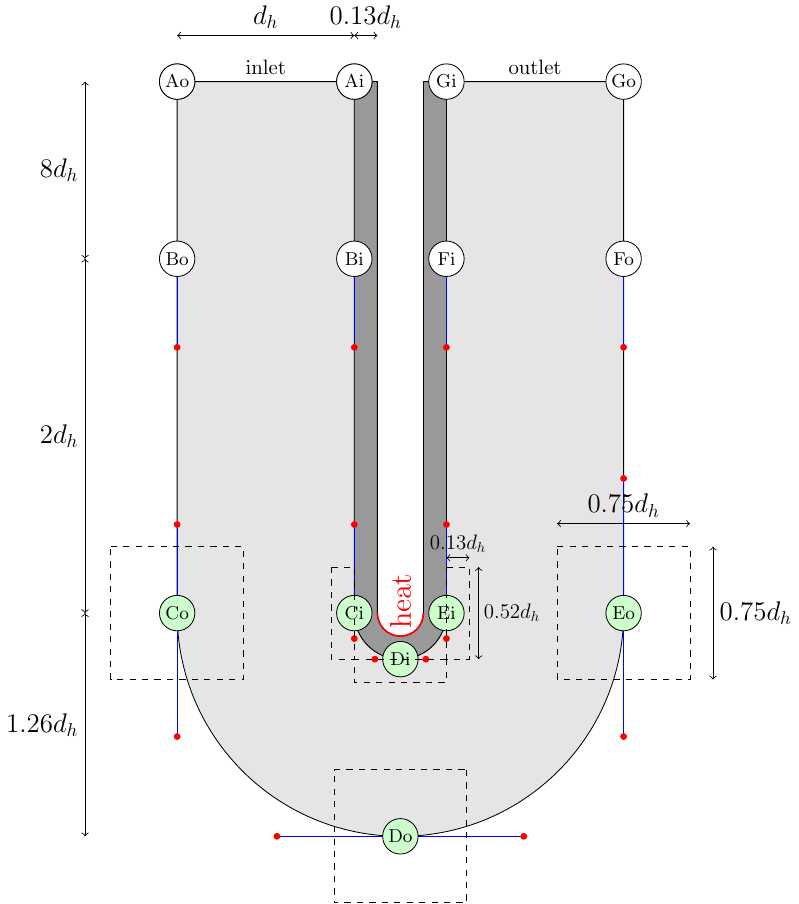}}
    }
	\caption{Parameterised geometry with boundary points (green) and curve parameters (red). The Figure is adapted and rotated from~\cite{DSB+22}}
    \label{fig:UseCase2Parameter}
    \vspace{-0.4cm}
\end{figure}

\subsection{Use Case 2: U-Bend Dataset}
\label{subsec:usecase2}
\jde{
The second use case is from the field of fluid dynamics. Flow deflections are a prevalent reason for energy dissipation in technical systems. Typical applications include large-scale piping grids or cooling channels of gas turbine blades. Fig.~\ref{fig:UseCase2Parameter} depicts the U-bend, which is described by 28 design parameters consisting of six boundary points (green) and 16 curve parameters (red). The dataset is publicly available~\cite{decke2023dataset} and the inclined reader is referred to the relevant paper for a detailed description. Overall, the amount of parameters characterizing the problem offers considerable flexibility in its design and potential for optimisation.

Utilizing computational fluid dynamics methods, the two target values---\textbf{Pressure Loss~$J_p$} and \textbf{Cooling Power~$J_T$}---are evaluated. The Navier-Stokes equations determine the velocities of the fluid and the pressure. In addition, the energy equation is used to consider the heat conduction and the convective heat transfer between fluid and solid using a multi-physics solver. Target $J_T$ is inversely defined, meaning that a lower value signifies higher cooling power. However, these two objectives inherently conflict, as no single design can optimise both simultaneously, presenting a classic multi-objective optimization challenge.

We utilize a subset of the data, focusing solely on the parameter representation of the dataset to maintain comparability with use case 1~\cite{decke2023dataset}. In contrast to use case 1 an identical and independent random sampling is used by the authors to generate the dataset. The training set comprises 800 designs, while the test set includes 200 designs. 
}
\section{Results}
\label{sec:results}
\dbo{
We illustrate the versatility of our pipeline using the presented use cases and structure our results in the subsequent subsections as follows: Subsection~\ref{subsec:reg_perf} showcases the performance of the involved ML-surrogate models in the prediction task. Subsection~\ref{subsec:results_xai} provides valuable insights into the regression task's outcomes. An evaluation and discussion of the results obtained with an evolutionary optimisation strategy are presented in Subsection~\ref{subsec:optimisation_results}. Moreover, in Subsection~\ref{subsec:validation}, we validate the acquired solution candidates by comparing the predictions of the surrogate models against the numerical simulations and finally present a performance criterion for selecting the final results.
}

    \begin{figure}[b!]
        \centering
         \includegraphics[width=.42\linewidth]{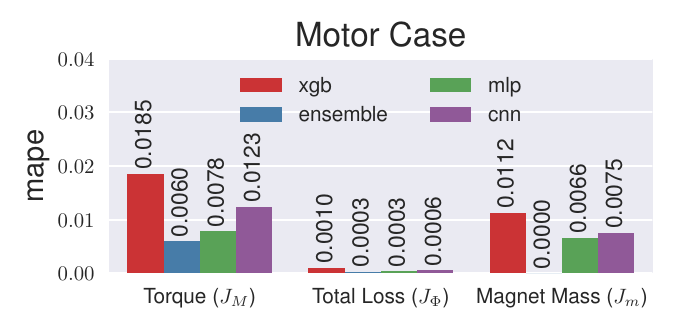}\includegraphics[width=.38\linewidth]{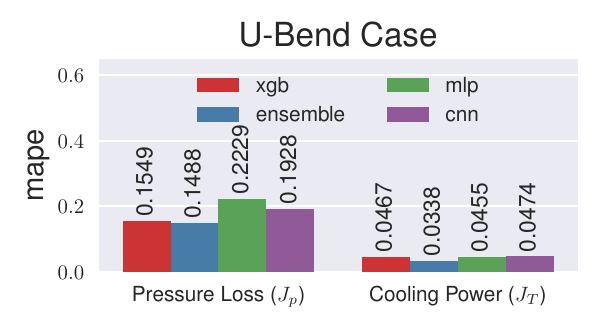}
        \includegraphics[width=.42\linewidth]{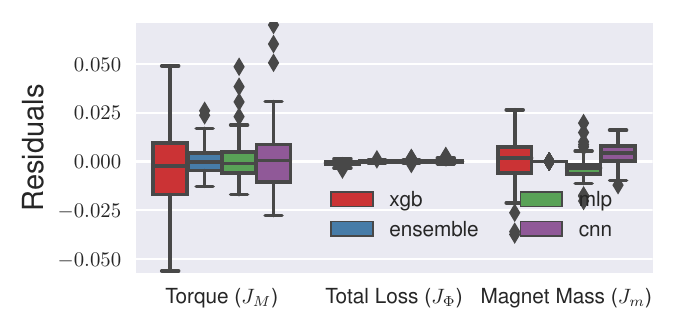}\includegraphics[width=.38\linewidth]{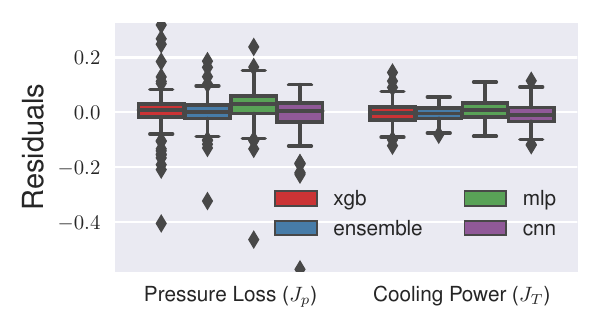}
        \caption{Assessment of surrogate model performance using MAPE scores and residual analysis on the test sets for each use case.}
        \label{fig:scores_all}
        \vspace{-0.4cm}
    \end{figure}

\subsection{Prediction Performance of Surrogate Models}
\label{subsec:reg_perf}
\dbo{

    The prediction performance of the surrogate models is evaluated using common regression metrics. We evaluate on the test dataset (20\%) the mean absolute percentage error (MAPE). This metric allows evaluation and comparison of the models' performance regardless of the target's scale, as can be seen in the top half of Fig.~\ref{fig:scores_all}. Here, MAPE values below 1.9\% for the \textbf{Torque~($J_m$)}, \textbf{Total Loss~($J_\Phi$)} and \textbf{Magnet Mass~($J_m$)} indicate good prediction performance on the Motor dataset. The U-Bend dataset poses a more difficult problem for the surrogate models, evident by the significantly higher MAPE scores using any surrogate model for predicting the target values \textbf{Pressure Loss~($J_P$)} and \textbf{Cooling Power~($J_T$)}. 

    The bottom part of Fig.~\ref{fig:scores_all} shows plots of the residuals, underscoring the presence of notable residual values for specific data points, particularly in predicting $J_T$ for the U-Bend dataset. These outliers likely contribute to the elevated MAPE values observed for the same target variable, given the sensitivity of this metric to extreme values. 
    }
\jde{
    Another potential explanation for the comparatively higher prediction error in the U-Bend case is the greater dimensionality of the input space, which has 13 dimensions more than the Motor case with 15 dimensions. Utilising the MSE during training and hyperparameter optimisation leads to a particularly severe penalty for outliers. Consequently, the optimisation process becomes more challenging in regions featuring low target values, as they have less importance when squared. Moreover, the U-Bend~case features an entirely turbulent flow, where minor parameter adjustments can trigger vortex formation and introduce discontinuities in the objective function. Additionally, the cases differ fundamentally in their physics: the Maxwell equation is a first-order linear partial differential equation, whereas the Navier-Stokes equation is a second-order nonlinear system with five solution variables, potentially accounting for the variance in prediction accuracy across our two use cases.
}

    \begin{figure}[b!]
        \centering
        \includegraphics[width=\linewidth]{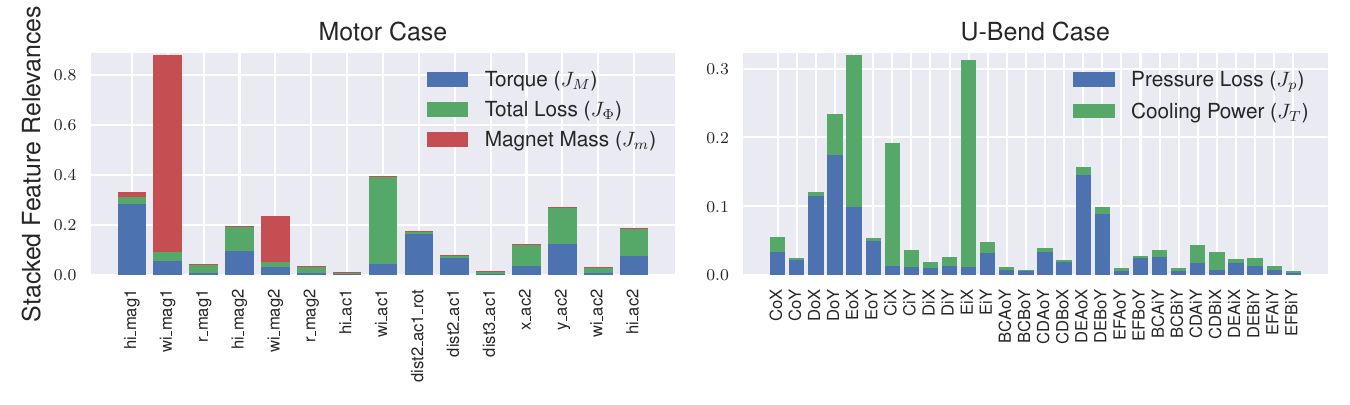}
        \caption{Analysis of feature importance. The plots show the stacked bars of feature important values obtained from the XGBoost models for each target value and use case.}
        \label{fig:relevances}
    \end{figure}

\subsection{Identifying Critical Features and relevant dependencies}
\label{subsec:results_xai}
\begin{figure}[t!]
        \centering
        \includegraphics[width=.64\linewidth]{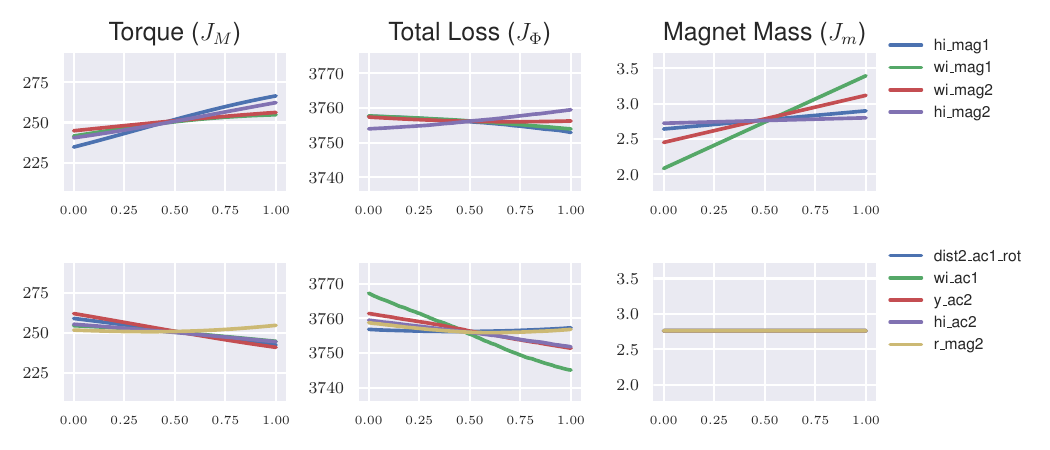}\includegraphics[width=.36\linewidth]{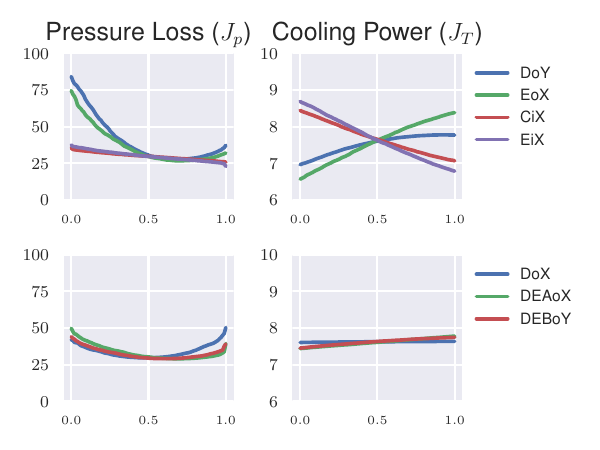}
        \caption{Partial dependency plots for selected features and target values for the Motor and the U-Bend datasets.}
        \label{fig:partial_dependencies}
    \end{figure}

\dbo{
    We start considering the evaluation of feature importances of each XGB model trained independently to predict a single output value in each of the presented use cases. 
    The corresponding importance values are shown in Fig.~\ref{fig:relevances}. In the Motor case, we can observe notable feature importances for the design parameters $wi\_mag1$, $wi\_mag2$ and $hi\_mag1$ concerning the output value \textbf{Magnet Mass~($J_m$)}. It was anticipated that these parameters would strongly correlate with the magnet's mass since it is a function of its size. 
    
    Regarding the output value \textbf{Total Loss~($J_\Phi$)} the high importance value for the design parameter $wi\_ac1$, corresponding to the width of one of the air cavities, can be explained by the fact, that air cavities influence the magnetic flux from the magnets to the air gap and could affect the overall loss and efficiency. Finally, the most relevant feature for the target \textbf{Torque~($J_M$)} is given by the parameter $hi\_mag1$. This correlates with the knowledge that the magnet components influence the intensity of the magnetic flux and, therefore, the torque. 
    
    The partial dependency plots presented in Fig.~\ref{fig:partial_dependencies} underscore the identified correlations using the feature importances. The curves show mainly linear dependencies for the motor case with high influence of the width parameters of the magnets and low influence of the air cavity parameters on the \textbf{Magnet Mass~$J_m$}. Conversely, the \textbf{Total Loss~$J_\Phi$} is primarily influenced by air cavity related parameters. and we can notice extremely low influence of air cavity parameters for the target value $J_m$.    
}

\jde{

    In the U-Bend case, the parameter DoY is crucial for \textbf{Pressure Loss~($J_P$)}, influencing the channel width at the outer layer's deflection point. This is expected, since altering the channel's width affects the local Reynolds number, i.e. the turbulence and vortex formation that contribute to $J_P$. Although EoX also impacts $J_P$, its effect is less pronounced. Outer curve parameters notably affect $J_P$, as optimal flow alignment can prevent vortex formation, whereas too small parameters may cause excessive deflection and stall.
    For \textbf{Cooling Power~$J_T$}, parameters EoX, CiX, and EiX are pivotal \dbo{with the highest importance values.} These parameters are especially relevant to the inner layer and directly impact the heated surface area, where a thinner surface lowers heat conduction resistance. Given the high Reynolds numbers, the dominant heat transfer mechanism from the solid to the fluid is convective, making a thin solid layer beneficial. Narrowing the channel---mainly managed by outer parameters---enhances cooling performance by promoting efficient flow contact.
}

\subsection{Evaluation of Solution Candidates in the Multiobjective Optimisation Task}
\label{subsec:optimisation_results}

\begin{figure}[b!]
        \centering
        \includegraphics[width=.5\linewidth]{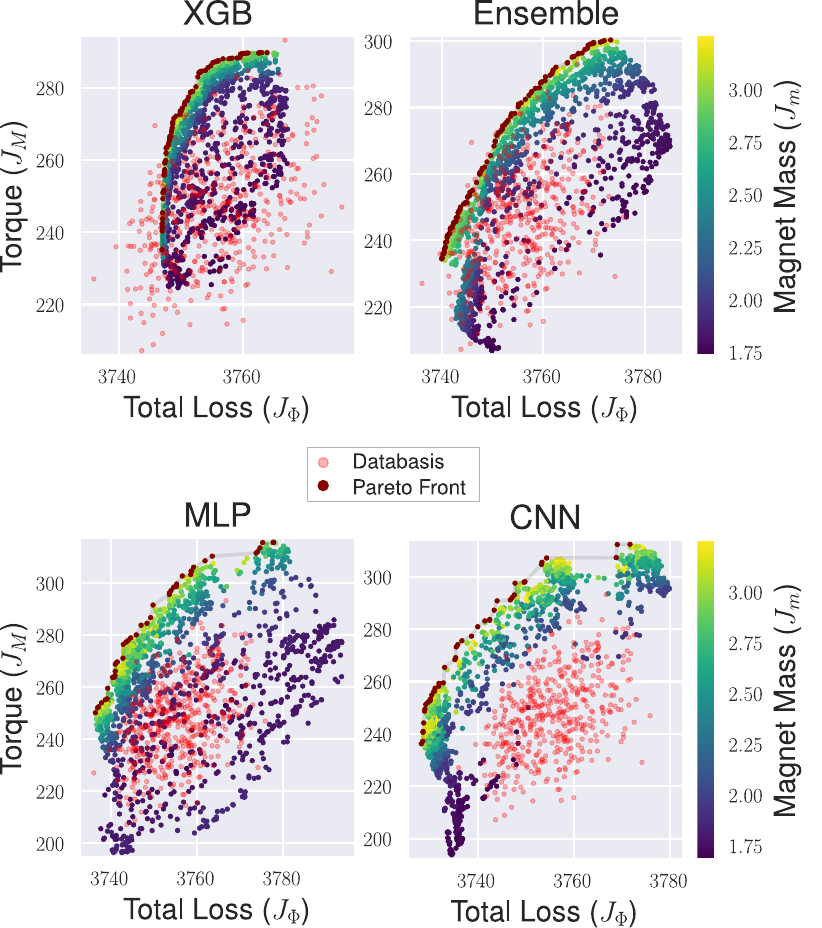}\includegraphics[width=.5\linewidth]{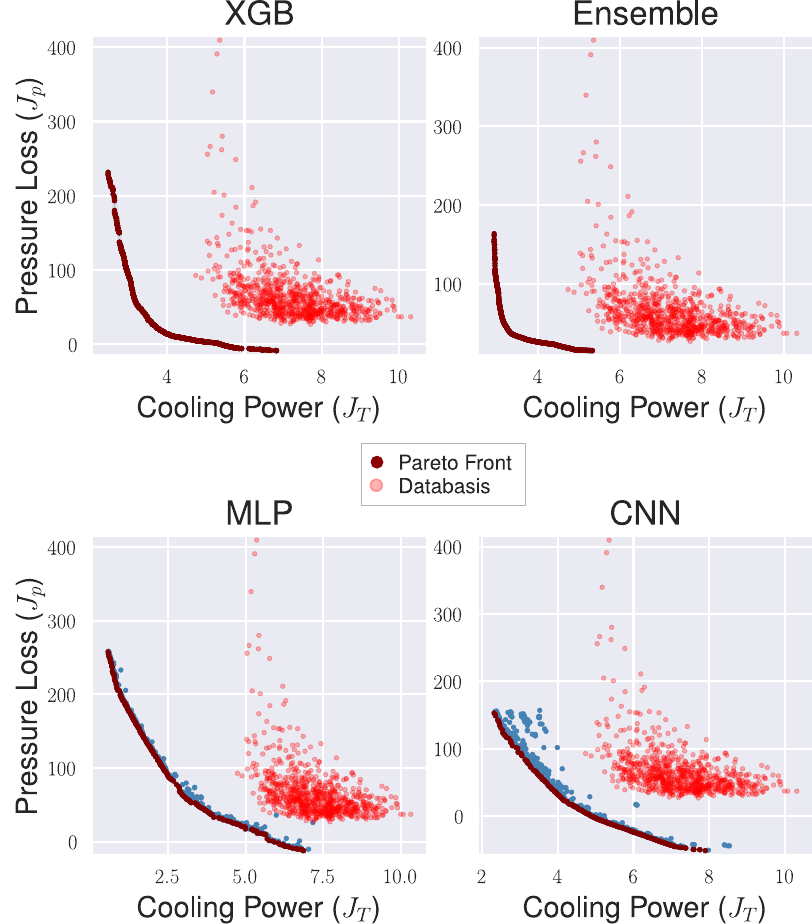}
        \caption{Optimisation Results for the Motor Use Case (left) and the U-Bend dataset (right) with EA in a combination of four different surrogate models}
        \label{fig:pareto_predictions}
    \end{figure}

\dbo{
    The optimisation results acquired with the four surrogate models in combination with the evolutionary algorithm for the Motor case are presented in Fig.~\ref{fig:pareto_predictions} on the left. With our presented EA optimisation strategy, we can evaluate around eight million points using the surrogate models instead of the numerical simulations. The red dots represent the data points available in the original database used for train and test of the surrogate models. The coloured dots correspond to the predictions of each surrogate model for the final population (2.000 samples) of points present in the last iteration of the evolutionary algorithm. The colour of the points represents the value of the \textbf{Magnet Mass~$J_m$}. The Pareto frontiers with dark red points are based on the model's prediction. We can notice that the frontiers given by the prediction from the MLP and CNN surrogates are far away from the original database. In contrast, the results with the XGB and Ensemble surrogate models deliver solution candidates which only outperform the database in regions of higher torque values ($J_M > 250$). 
}

\jde{

    In the U-Bend case, only two objectives necessitate a two-dimensional Pareto frontier, as depicted in Fig.~\ref{fig:final_validation_pfronts}. Evolutionary optimization produces Pareto frontiers for each model, presenting design candidates superior to those in the initial database, shown as grey dots. However, a critical observation is that three of the four models predict physically unrealistic negative pressure loss values, suggesting a lack of understanding of the physical principles involved. Recalculating the design candidates from these frontiers will further assess the models' performance.
}

\subsection{Validation of Solution Candidates in the Multiobjective Optimisation}
\label{subsec:validation}
\dbo{
    In the last stage of our evaluation strategy, we compare the predictions of the surrogate models, i.e. the selected solutions at the Pareto frontiers, against the true values using the original physical numerical models. The first indicator for the surrogate models' quality in predicting target values for the presented use cases is reflected in Table~\ref{tab:pareto_performance_indicators} through the simulation rate. A higher simulation rate reveals the models' ability to capture parameter interrelationships and the ability to generate many plausible solutions.
    
     \begin{figure}[h!]
        \centering    
        \includegraphics[width=.55\linewidth]{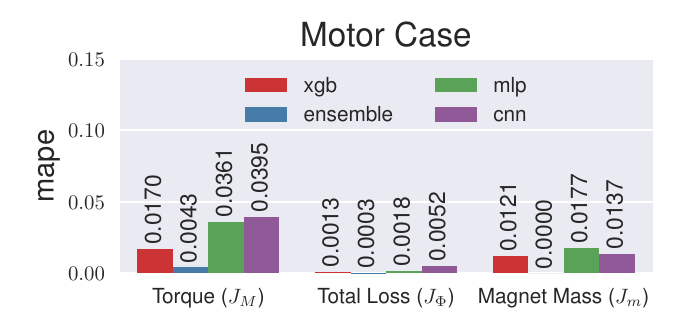}\includegraphics[width=.42\linewidth]{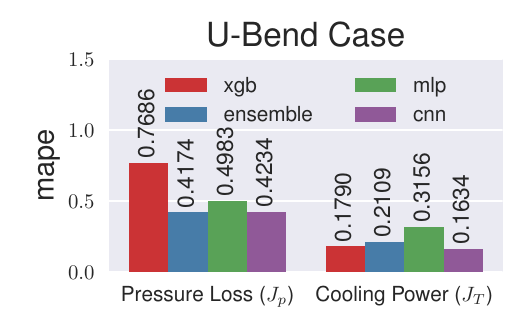}
        \caption{MAPE of the final validation of the Motor and U-Bend Use Cases}
        \label{fig:final_validation_mape}
        \vspace{-0.4cm}
    \end{figure}

         \begin{figure}[b!]
        \centering    
        \includegraphics[width=.5\linewidth]{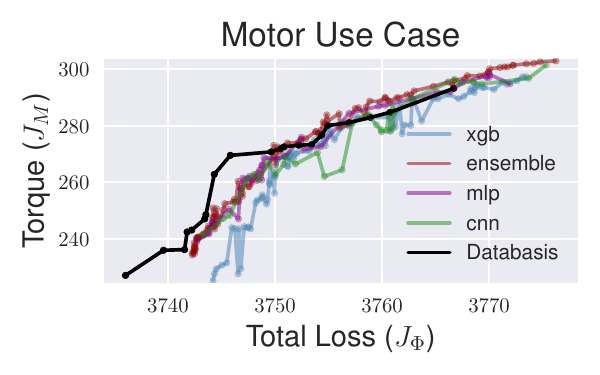}\includegraphics[width=.5\linewidth]{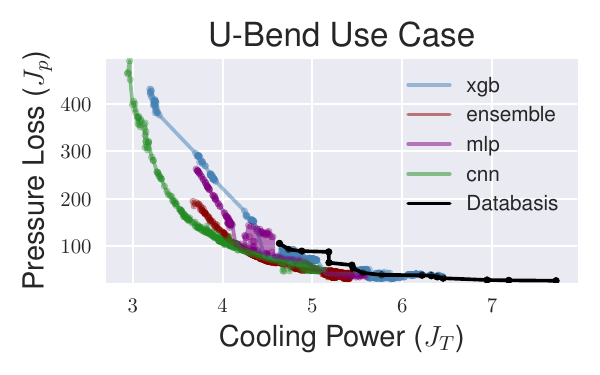}
        \caption{Pareto frontiers of the final validation of the Motor and U-Bend Use Cases}
        \label{fig:final_validation_pfronts}
        \vspace{-0.4cm}
    \end{figure}
    
    In the motor use case, we achieve a 100\% simulation rate with the ensemble model and---due to the complexity of the U-Bend case---only up to 78\% of the solution candidates provided likewise with the ensemble were plausible. Next, in Fig.~\ref{fig:final_validation_mape} we validate the solution candidates using the numerical simulations and evaluate the results with the MAPE score. We generally appreciate higher MAPE scores for the U-Bend case reflecting higher complexity in the design space and more challenging physical principles. However, the deviation of model predictions in areas of interest in the optimisation task is mainly affected by the lower presence of samples in the training database. If the simulation rate is extremely low, an additional training cycle of the surrogate models using the simulated data points i.e., plausible samples from the Pareto frontiers, should be considered. These additional data for training can provide additional information about regions of lower density and closer to the Pareto frontiers to the surrogate models.

    The numerical validation and final results of the selected points are shown in Fig.~\ref{fig:final_validation_pfronts}. The optimisation task was successful in both cases. In the motor use case, we appreciate better solutions candidates at higher Torque values as appreciated in the predictions in subsection~\ref{subsec:optimisation_results}. Particularly in the U-Bend case, the simulation results are closer to the database and reflect higher deviations from the predicted values in Fig.~\ref{fig:pareto_predictions} on the right. However, the obtained solutions candidates outperform the original database Pareto frontier in all regions considering the two objective values $J_T$ and $J_P$. These results emphasise the versatility of our strategy and set the focus on the quality of the final solution candidates, which do not depend on the prediction accuracy of the surrogate models. 

\begin{table}[h!]
    \centering
    \caption{Performance indicators of validated solutions candidates concerning the original database (DB) for both use cases using Generational Distance (GD), GD-Plus (GD+)~\cite{ishibuchi_modified_2015} and the Hypervolume Indicator (HV)~\cite{fonseca_improved_2006}.}
    \vspace{0.2cm}
    \begin{tabular}{lccccc|ccccc}
    \cmidrule{2-11}
    {} & \multicolumn{5}{c}{ Motor Case }  & \multicolumn{5}{|c}{ U-Bend Case }\\
    \cmidrule{2-11}
    {} & Sim. Rate &  GD &  GD+ & HV$\uparrow$  & & & Sim. Rate & GD &   GD+ &  HV$\uparrow$ \\
    \midrule
    Database (DB)   & -        & - & - & 1.753                  & & & -      & - & - & 0.973         \\
    DB + XGB        & 98.91\%  & 0.059 & 0.012 & 1.766          & & & 69.2\% & 0.063 & 0.033 & 1.077  \\
    DB + Ens.       & 100\%    & 0.048 & 0.018 & \textbf{1.795} & & & 78.3\% & 0.062 & 0.009 & 1.112  \\
    DB + MLP        & 80.43\%  & 0.038 & 0.010 & 1.774          & & & 20.9\% & 0.111 & 0.078 & 1.091  \\
    DB + CNN        & 91.81\%  & 0.041 & 0.012 & 1.770          & & & 53.6\% & 0.128 & 0.064 & \textbf{1.157}  \\
    \bottomrule
    \end{tabular}
    \label{tab:pareto_performance_indicators}
\end{table}

    In our presented experiments we used four surrogate strategies, which can be visually compared using the Pareto frontiers. Anyway, adding additional surrogates and combining them with additional optimisation strategies could become unmanageable to qualitatively compare. We propose using performance indicators as shown in Table~\ref{tab:pareto_performance_indicators}. This allows a quantitative comparison of the validated Pareto frontiers and the filtering of only the most significant results. In the motor use case, we see higher performance of the results with the ensemble strategy with a Hypervolume (HV) of 1.795 and in the U-Bend case we observe the best results with the CNN surrogate and an HV of 1.157. The values of the HV are calculated using normalized target values and reference points of the maximal values present in the original databases. 
}

\section{Conclusion}
\label{sec:conclusion}

\dbo{
Our investigation underscores the efficacy of employing diverse machine learning surrogate models for predicting output values across heterogeneous engineering use cases. In particular, our analysis reveals that although the Motor dataset exhibits commendable performance with MAPE values below 1.9\%, the U-Bend dataset poses greater challenges, evident in higher MAPE values. At the hand of a feature importance analysis and partial dependency plots, we determined key parameters significantly influencing output values within both datasets. Validating the solution candidates against validated values using numerical simulations confirms the efficacy of the surrogate models. Despite higher complexity in the U-Bend case with some deviations attributed to the scarcity of training data in certain regions, our approach achieves efficient results. It showcases the versatility of our pipeline in addressing diverse optimization tasks and the evaluation strategy can be applied in a general way in future work to multiple use cases. Future investigations will focus on iteratively refining the optimization cycle and expanding the dataset to enhance model performance, the robustness of our methodology and its applicability in addressing complex engineering problems.
}

\bibliographystyle{splncs04}
\bibliography{bib/db}
\end{document}